\documentclass[journal]{IEEEtran}
\newcommand{\RNum}[1]{\uppercase\expandafter{\romannumeral #1\relax}}
\usepackage{mathrsfs}
\usepackage{amssymb}
\usepackage[mathcal]{euscript}
\usepackage{diagbox}
\usepackage{cite}
\usepackage[T1]{fontenc}
\usepackage{graphicx}
\usepackage{CJKutf8}
\usepackage{makecell}
\usepackage{psfrag}
\usepackage{url}
\usepackage{stfloats}
\usepackage{amsmath}
\usepackage{array}
\usepackage{float}
\usepackage{multirow} 
\usepackage{hyperref}
\usepackage{amsthm}
\usepackage{color}
\usepackage{multicol}
\usepackage{stfloats}
\usepackage{enumerate}
\usepackage{subfigure}
\usepackage{booktabs}  %±í¸ñ

\allowdisplaybreaks[4]
\usepackage[ruled]{algorithm2e}
\usepackage{algorithmic} %format of the algorithm
\usepackage{xurl}
\usepackage{xcolor}

\ifCLASSINFOpdf
\else
\fi
\begin{document}
\begin{CJK}{UTF8}{gbsn}
\title{Resource-Aware LLM Reasoning for Mobile Edge General Intelligence}

% \title{Streaming AIGC in Wireless Communication Networks: A Quantum Machine Learning Approach }

%AI-empowered Adaptive Coverage Enhancement in 6G: Opportunities and Challenges

%\author{\IEEEauthorblockN{Michael Shell}
%\IEEEauthorblockA{School of Electrical and\\Computer Engineering\\
%Georgia Institute of Technology\\
%Atlanta, Georgia 30332--0250\\
%Email: http://www.michaelshell.org/contact.html}
%\and
%\IEEEauthorblockN{Homer Simpson}
%\IEEEauthorblockA{Twentieth Century Fox\\
%\and
%\IEEEauthorblockN{James Kirk\\ and Montgomery Scott}
%\IEEEauthorblockA{Starfleet Academy\\
%San Francisco, California 96678-2391\\
%\author{Ruichen Zhang$^*$ Ke Xiong$^*$, Wei Guo$^*$ \\
%\small
%$^*$School of Computer and Information Technology, Beijing Jiaotong University, Beijing 100044, P. R. China, \\
%Beijing 100044, China\\
%E-mail:18120449@bjtu.edu.cn}
	\author{Mingyi Luo, Ruichen Zhang, Xiangwang Hou, Jun Du, \\ Chunxiao Jiang,~\IEEEmembership{Fellow,~IEEE}, Yong Ren, and Shiwen Mao,~\IEEEmembership{Fellow,~IEEE}

\thanks{M. Luo is with Tsinghua Shenzhen International Graduate School, Tsinghua University, Shenzhen (e-mail: luomy24@mails.tsinghua.edu.cn).}
\thanks{R. Zhang is with the College of Computing and Data Science, Nanyang Technological University, Singapore (e-mail: ruichen.zhang@ntu.edu.sg).}
\thanks{X. Hou (corresponding author) is with the Department of Electronic Engineering, Tsinghua University, Beijing 100084, China. (e-mail: xiangwanghou@163.com).}
\thanks{J. Du and Y. Ren are with the Department of Electronic Engineering and also with
the State Key Laboratory of Space Network and Communications, Tsinghua University, Beijing 100084, China (e-mail: jundu@tsinghua.edu.cn, reny@tsinghua.edu.cn).}
\thanks{C. Jiang is with the Beijing National Research Center for Information
Science and Technology, and the State Key Laboratory of Space Network
and Communications, Tsinghua University, Beijing 100084, China (e-mail:
jchx@tsinghua.edu.cn).}
\thanks{S. Mao is with the Department of Electrical and Computer Engineering, Auburn University, Auburn, USA (e-mail: smao@ieee.org).}
}
\maketitle

\begin{abstract}
The rapid advancement of large language models (LLMs) has enabled an emergence of agentic artificial intelligence (AI) with powerful reasoning and autonomous decision-making capabilities. This integration with edge computing has led to the development of Mobile Edge General Intelligence (MEGI), which brings real-time, privacy-preserving reasoning to the network edge. However, deploying LLM-based agentic AI reasoning in MEGI environments poses significant challenges due to the high computational demands of reasoning and the limited resources of edge devices. To address these challenges, we propose a joint optimization framework for efficient LLM reasoning deployment in MEGI. First, we systematically review enhancement methods to identify mechanisms suitable for edge adaptation. Subsequently, we present a distributed framework that synergizes reasoning enhancement via adaptive CoT prompting with scalable deployment through a distributed MoE architecture. An important innovation of this approach involves modeling reasoning depth as a dynamic network resource variable, which is optimized jointly with expert activation and transmission power. This mechanism allows the system to dynamically regulate expert networks and reasoning complexity according to task requirements and device capabilities. Experimental evaluations in mobile edge environments demonstrate that the proposed framework effectively balances reasoning quality and resource efficiency. The results show that with less than one second of additional inference time, both accuracy and latency satisfaction rate can reach 90\%, validating the practical viability of deploying sophisticated LLM reasoning in resource-constrained MEGI systems.
\end{abstract}

\section{Introduction}
{
Recent advances in artificial intelligence (AI) demonstrate that reasoning is fundamental to intelligence. The recently proposed DeepSeek-R1 model has demonstrated superior performance in mathematics and coding competitions by employing sophisticated reasoning strategies such as self-verification, reflection, and adaptive decision-making \cite{guo2025deepseek2}. This capability emphasizes a paradigm shift toward Agentic AI, where large language models (LLMs) evolve from passive text generation to autonomous, goal-directed behavior. By integrating planning, memory, and tool usage, Agentic AI dynamically decomposes tasks and formulates multi-step strategies to navigate evolving environments \cite{10679152,11049053}.
Deploying autonomous agents leads to an introduction of Mobile Edge General Intelligence (MEGI) \cite{zhang2025toward}. While cloud-centric solutions such as ChatGPT \cite{achiam2023gpt} offer substantial computational capacity, they often fail to meet the strict requirements of real-time tasks, including automated fault detection and privacy-sensitive interactions. In such scenarios, the propagation latency and data sovereignty risks associated with cloud offloading become prohibitive. MEGI mitigates these limitations by instantiating reasoning capabilities directly at the network edge, thereby ensuring service continuity and privacy preservation.

However, deploying LLM-based agentic AI systems with reasoning capability in MEGI environments introduces significant technical challenges\cite{10879580}. Edge devices operate under tight resource constraints, including limited computational power, memory capacity, and energy budgets. Furthermore, the sophisticated reasoning mechanisms of agentic AI impose substantial computational demands. The dynamic nature of mobile environments introduces additional complexity through variable network conditions, device mobility, and heterogeneous hardware capabilities.  Existing optimization strategies frequently struggle to reconcile these constraints with the demands of high-quality reasoning \cite{feng2024optimizing,husom2025sustainable,wang2022self,yi2025edgemoe}. Although recent studies have addressed specific aspects of edge deployment, such as service offloading \cite{feng2024optimizing}, model quantization \cite{husom2025sustainable}, or sparse activation architectures like EdgeMoE \cite{yi2025edgemoe}, they typically view the inference process as a static workload. This oversight fails to account for the dynamic characteristics of reasoning-enhanced tasks, where inference quality fluctuates with reasoning depth, and communication overhead scales with intermediate token transmission. Consequently, static methods are ill-suited for scenarios requiring flexible computation allocation among distributed experts. For instance, aggressive quantization is often employed to fit models onto constrained GPUs, yet this approach can degrade accuracy by up to 40\% compared to higher-precision variants \cite{husom2025sustainable}.

To address these challenges, this paper proposes a joint optimization framework that enables efficient deployment of LLM-based agentic AI systems with reasoning in MEGI environments. Rather than simply combining existing techniques, our framework establishes a theoretical link between reasoning quality and network resources. The framework enhances reasoning via adaptive Chain-of-Thought (CoT) prompting and ensures scalability through a distributed Mixture of Experts (MoE) architecture. Specifically, CoT prompting decomposes complex problems into manageable steps, allowing the system to adjust reasoning depth based on task complexity and resource availability. The distributed MoE architecture allocates tasks across edge devices, maintaining high reasoning quality while achieving computational scalability.
% Through comprehensive analysis and experimental validation, this work establishes a foundation for reasoning-enabled AI in wireless communication systems. It provides methodologies for implementing and evaluating reasoning mechanisms in resource-constrained environments, and develops optimization strategies for diverse edge computing scenarios. 
The contributions of this work can be summarized as follows:
\begin{itemize}
\item We provide an overview of methods for enhancing LLM agentic AI reasoning capabilities across different phases of model development, including pre-training (e.g., model scaling and improved model architecture), fine-tuning (e.g., Supervised Fine-Tuning (SFT), Reinforcement Learning from Human Feedback (RLHF)), and inference (e.g., CoT prompting and self-consistency). This analysis highlights technical prerequisites for adapting reasoning-capable agents to edge environments.
\item We propose a joint optimization framework for LLM agentic AI reasoning in MEGI, featuring a distributed MoE architecture integrated with CoT prompting. A key distinction is our reasoning-aware resource management mechanism. We theoretically model the trade-off between reasoning depth and energy consumption, enabling the systematic minimization of total system energy by jointly optimizing token assignment, transmission power, and adaptive reasoning depth.
\item We conduct extensive experimental evaluations, ranging from local deployment validation to system-level performance assessment. The results verify the effectiveness of the reasoning enhancement techniques and demonstrate the framework's ability to balance reasoning quality with resource efficiency. These findings provide empirical evidence supporting the practical viability of sophisticated LLM reasoning in MEGI environments.
\end{itemize}

}

\section{Overview of LLM Reasoning For MEGI}
\subsection{Definition of Reasoning}
Reasoning in the context of LLMs refers to the cognitive ability to perform systematic logical inference, problem decomposition, and multi-step analytical thinking to reach solutions\cite{guo2025deepseek2}. Unlike simple pattern matching or direct retrieval from training data, reasoning involves the manipulation of abstract concepts, establishment of causal relationships, and generation of novel solutions through the systematic combination of existing knowledge. This capability allows LLMs to address complex problems requiring sequential and strategic reasoning.
% \cite{huang-chang-2023-towards}
Recent advances in agentic AI have pushed LLMs beyond passive text generation toward autonomous, goal-driven behavior. LLMs can plan, act, and adapt within dynamic environments using conditional logic, memory, and tool use. In such systems, reasoning forms the decision-making core, guiding step-by-step planning and execution in response to changing conditions.
For example, Fu et al. \cite{10495699} demonstrated the effectiveness of LLM reasoning in autonomous driving by employing CoT and ReAct strategies with GPT-3.5. Their system achieved over 60\% zero-shot pass rate\footnote{The zero-shot pass rate refers to as the percentage of tasks successfully completed without any task-specific training.} in a highway simulation environment, significantly outperforming
reinforcement learning methods that require extensive iterations to reach similar performance levels. 

In the following sections, we review and categorize existing methods for enhancing LLM reasoning capabilities across three phases of agentic AI pipeline.

% However, in contrast to the application of explicit human logic rules, the reasoning capability of LLM is significantly influenced by the distribution of training data and the scale of model parameters. This renders LLM susceptible to inherent challenges, such as poor interpretability and logical jumps.

% These inherent challenges limit the effective application of LLM reasoning capabilities in real-world scenarios, especially in resource-constrained MEC environments. In order to overcome these limitations, researchers have investigated a range of methods to augment LLM reasoning capabilities across three distinct phases: pre-training, fine-tuning, and inference time. These methods have achieved some success in enhancing LLM reasoning ability from different perspectives, such as model construction, parameter optimization, and reasoning strategy, respectively, but there are still limitations in mobile edge computing scenarios.

\subsection{LLM Reasoning Methods}
Enhancing the reasoning abilities of LLMs has been approached through various methods across different phases of model development.

\begin{figure*}[t]
\centering{\includegraphics[width=0.93\textwidth]{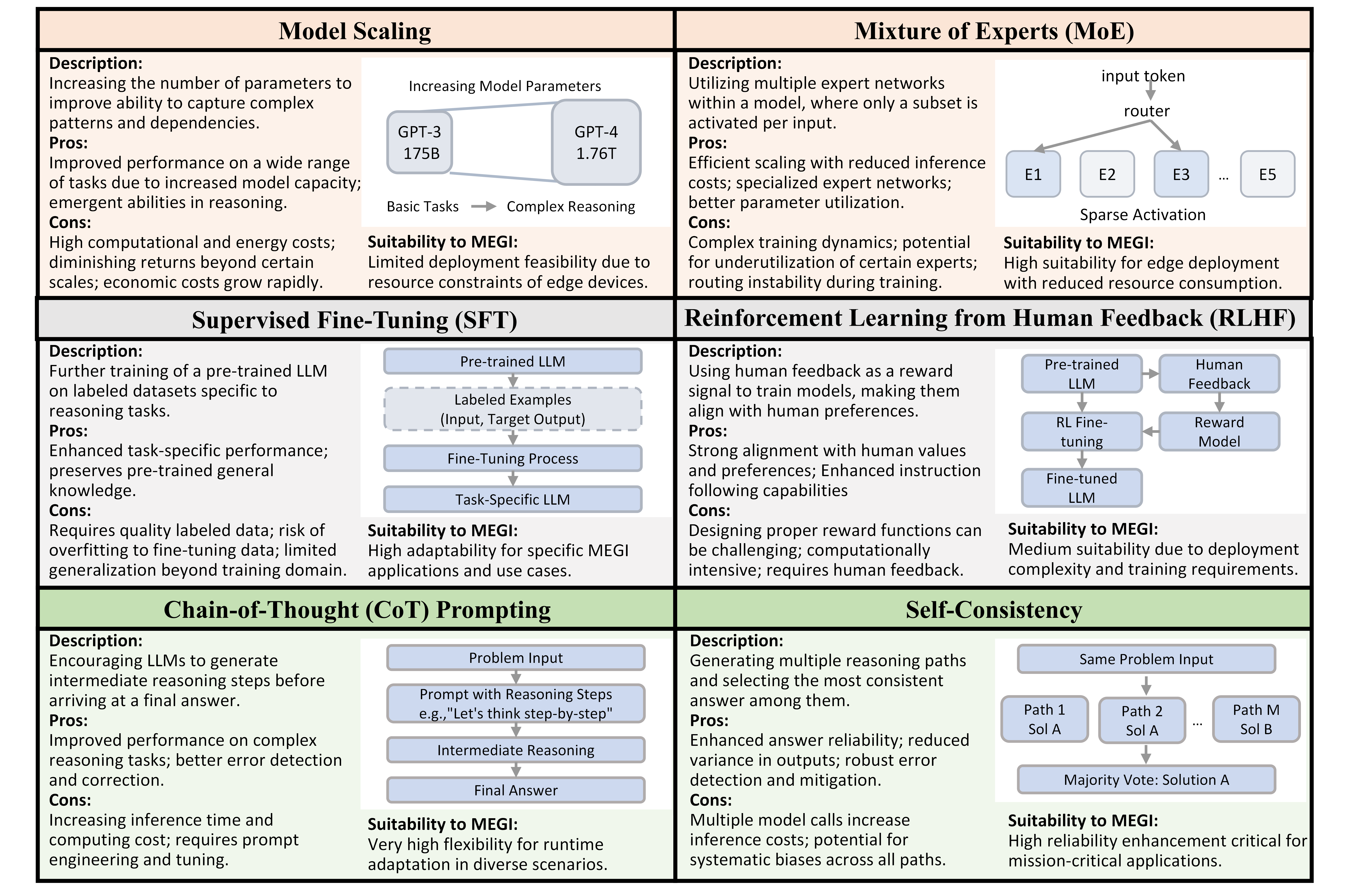}}
\caption{Comparison of methods for enhancing LLM reasoning capabilities across different phases of model development\cite{xu2025towards}. The diagram categorizes approaches into three phases: Pre-Training (e.g., Model Scaling and MoE), fine-tuning (e.g., SFT and RLHF), and Inference (e.g., CoT Prompting and Self-Consistency).For each method, the corresponding workflow, advantages, and limitations are illustrated.}\label{Fig1}
\end{figure*}

\subsubsection{Pre-Training Phase}
The pre-training phase establishes foundational reasoning capabilities through architecture and large-scale data, determining the model’s base capacity for recognition and logical inference.

\begin{itemize}
    \item \textbf{Model Scaling:} Model scaling refers to increasing model size by expanding parameters, layers, or architecture dimensions. This approach establishes robust foundational reasoning capabilities, with larger models demonstrating emergent reasoning behaviors, particularly in complex multi-step problems requiring extensive knowledge integration. 
    For instance, the progression from GPT-3 (175 billion parameters) to GPT-4 (estimated 1.76 trillion parameters) demonstrates significant improvements in reasoning performance. Empirical benchmarks, such as the GSM8K mathematical reasoning dataset, show GPT-4 achieving 92\% accuracy, a substantial increase from GPT-3.5's 57\% \cite{achiam2023gpt}.

    % Larger Model Size: Increasing parameters (e.g., GPT-3: 175B → GPT-4: 1.7T) to capture complex patterns. \cite{kaplan2020scaling}
    % Limitation: High computational costs and deployment barriers.
    % However, this comes at the cost of high computational costs and deployment barriers. Training and deploying such large models require massive computational resources and energy consumption. The increased model size also poses challenges for deployment in resource-constrained environments, such as edge devices, due to limitations in memory and processing power.

    \item \textbf{New Model Architectures:} Emerging designs such as Mixture of Experts (MoE) and Multi-Agent systems aim to enhance reasoning while maintaining computational efficiency. MoE frameworks utilize sparse activation patterns where only specialized expert networks process each input, while Multi-Agent architectures enable collaborative reasoning through distributed intelligent agents specializing in different problem aspects. These new model architectures are crucial for balancing reasoning capability with computational efficiency, particularly in resource-constrained environments.
    For example, Yi et al. \cite{yi2025edgemoe} developed EdgeMoE, a sparse architecture designed for mobile device deployment. By selectively activating relevant expert networks, it achieves enhanced reasoning with 5\% to 18\% reduced memory usage and 1.2 to 2.7 times faster inference, enabling real-time performance for models exceeding 10 billion parameters.
    
    % Sparse activation (e.g., DeepSeek-MoE (1.7B parameters with sparse experts) balances scale and efficiency.).\cite{shazeer2017outrageously}
    % The MoE approach allows the model to distribute its computational load, making it more efficient and suitable for resource-constrained environments. 
    % By engaging only a fraction of the model’s parameters during inference, the method greatly reduces the processing demand and lowers the active memory footprint, making large-scale reasoning more feasible on edge hardware, allowing for dynamic adjustment depending on task complexity and available resources.
\end{itemize}

\subsubsection{Fine-Tuning Phase} \label{section:fine-tuning}
The fine-tuning phase focuses on refining and specializing reasoning abilities without the computational expense of retraining models from scratch, allowing for targeted improvements and adaptation to specific domains.	

\begin{itemize}
    \item \textbf{Supervised Fine-Tuning (SFT):} SFT involves training pre-trained LLMs on domain-specific datasets with explicit input-output pairs, enhancing specialized reasoning and performance without the computational cost of training from scratch.
    For example, He et al. \cite{10591707} proposed an active inference approach for LLM inference offloading in cloud-edge computing environments. By fine-tuning on offloading decision datasets, the model learns to reason about resource allocation strategies between mobile devices and edge servers. Experiments show that this approach outperforms traditional offloading strategies, reducing latency by 20\% and achieving a 99\% task completion rate.
    
    % If the data is of high quality and contains a sufficient number of reasoning examples, the model's ability to reason in this domain is significantly enhanced.\cite{ouyang2022training}
    % Collecting data for SFT takes a lot of time and money.

    \item \textbf{Reinforcement Learning from Human Feedback (RLHF):} RLHF trains LLMs to enhance complex reasoning using reward-based learning. The model learns to solve problems by receiving feedback on the quality of its reasoning processes, enabling it to discover optimal reasoning strategies through trial and refinement.
    DeepSeek \cite{guo2025deepseek2} demonstrated the effectiveness of RLHF in enhancing LLM reasoning through their DeepSeek-R1-Zero model, which achieved significant improvements without requiring SFT. On the AIME 2024 mathematical reasoning benchmark\footnote{\url{https://maa.org/math
-competitions/american-invitational-mathematics-examination-aime/}}, their model's accuracy increased from 15.6\% to 71.0\%. 
    % Their approach utilizes Group Relative Policy Optimization (GRPO) to train the model through reward-based learning, where the model receives feedback on the quality of its reasoning processes and learns to generate more accurate solutions through trial and refinement. \cite{ziegler2019fine}

    % GRPO DPO \cite{rafailov2023direct} PPO \cite{schulman2017proximal}

    % However, designing effective reward functions is non - trivial. The rewards need to accurately reflect the desired reasoning outcomes and guide the model towards correct reasoning paths. Poorly designed rewards can lead to sub - optimal or even harmful model behaviors.
\end{itemize}

\subsubsection{Inference Phase}
The inference phase enhances reasoning capabilities after model deployment without modifying the underlying model parameters, providing dynamic reasoning improvements that can be adapted to specific problems and computational constraints.	

\begin{itemize}
    \item \textbf{Chain-of-Thought (CoT) Prompting:} CoT prompting enables LLMs to generate intermediate reasoning steps when solving complex problems, breaking complex tasks into sequential operations. This allows models to simulate human-like deliberation and provides transparency that facilitates the identification and correction of reasoning errors. Extensions of CoT include Tree-of-Thought, which explores multiple reasoning paths in a tree structure, and Graph-of-Thought, which models reasoning as a graph to support flexible, non-linear inference.
    For example, Feng et al. \cite{feng2024optimizing} proposed a CoT framework for optimizing microservice deployment in edge computing. By structuring decisions into intermediate steps, the model effectively manages trade-offs in latency, resource constraints, and service dependencies. Experimental results showed 8.3\% latency improvement over traditional methods while supporting real-time decisions.

    % For instance, in the process of solving a mathematical problem, the model first articulates each calculation or assumption before arriving at the final answer.  \cite{wei2022chain}

    % However, CoT prompting is scale-dependent. It works best for models with over 100 billion parameters. Smaller models may generate illogical chains, leading to decreased accuracy. Additionally, generating reasoning steps increases the token count, resulting in higher memory and computational costs during inference.

    \item \textbf{Self-Consistency:} Self-consistency generates multiple independent reasoning paths for the same problem and selects the most frequent answer through majority voting. It can be implemented by sampling diverse outputs from the model with different decoding randomness. This technique is essential for addressing inherent variability and potential errors in LLM reasoning, providing a robust validation mechanism.
    For example, Wang et al. \cite{wang2022self} demonstrated that self-consistency improves reasoning accuracy across diverse problem domains, achieving 17\% improvement on mathematical reasoning tasks and 11\% improvement on commonsense reasoning benchmarks. 

    % \item Self-Consistency: Self-consistency is a decoding strategy that improves reasoning reliability by generating multiple reasoning paths for the same problem and selecting the most frequent answer through majority voting. This approach leverages the principle that correct reasoning should be more consistent across different solution attempts. This approach can increase accuracy and reduce errors by leveraging the model's inherent variability in generating responses \cite{wang2022self}.

    % % However, it demands higher computational resources as multiple forward passes through the model are required. This can be prohibitive in scenarios with limited computational power or strict latency requirements.

    % \item Self-refinement: Self-refinement is an iterative improvement process where LLMs generate initial solutions, provide self-critique, and then produce improved versions based on their own feedback. This approach mimics human revision processes and can lead to higher-quality reasoning outputs \cite{madaan2023self}.
    % Intrinsic self-refinement provides feedback based solely on internal knowledge and reasoning. Extrinsic self-refinement, such as Program- Aided Language Models (PAL) or tool-calling, leverages external tools or programs to assist in refining the model's output. 
    
    % However, this approach adds complexity and potential latency due to the need for interaction with external systems.
\end{itemize}

In summary, Fig. \ref{Fig1} highlights that while existing methods improve reasoning, they exhibit distinct trade-offs: pre-training offers intrinsic capacity but high cost; fine-tuning provides specialization but requires data; and inference techniques enhance precision at the expense of latency. By systematically listing these pros and cons, Fig. \ref{Fig1} highlights that existing methods are typically studied in isolation and assume sufficient computational resources. When deployed in MEGI environments, their effectiveness may diminish due to energy, latency, and hardware limitations. Therefore, a joint optimization framework is essential to sustain reasoning quality while ensuring practical efficiency in edge deployment.

\section{A Joint Optimization Framework for LLM reasoning for MEGI}   \label{section:system_model}
\begin{figure*}[t]
\centering{\includegraphics[width=0.98\textwidth]{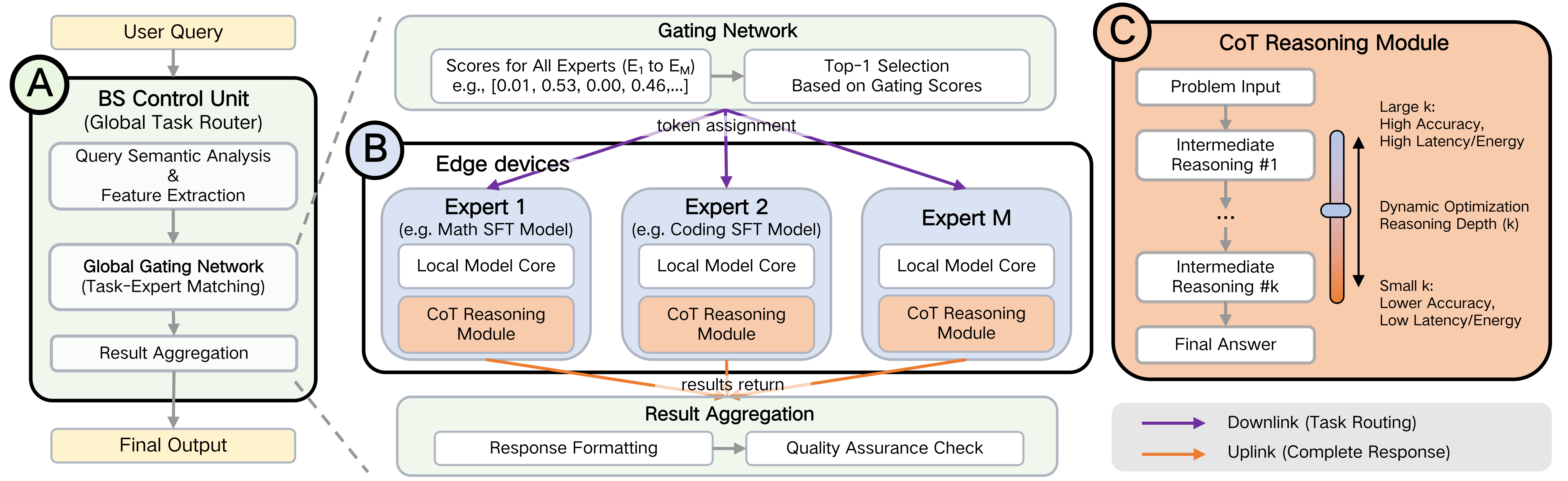}}
\caption{An overview of the proposed joint optimization framework for LLM reasoning for MEGI. The architecture consists of three main components: (A) BS Control Unit, (B) distributed edge devices hosting MoE-based expert networks, and (C) integrated CoT reasoning modules. It illustrates the end-to-end inference workflow, including expert selection, token assignment, parallel edge inference, and result aggregation.}
\label{Fig2}
\end{figure*} 
\subsection{Distributed MoE Architecture for MEGI}
\subsubsection{Challenge and Motivation}
Integrating LLMs into MEGI environments presents significant challenges. Edge devices operate under strict resource constraints, limiting their computational power and memory for deploying large-scale LLMs with reasoning capabilities. Additionally, the inference process of LLMs is characterized by high energy consumption, leading to rapid battery depletion in mobile devices. The dynamic nature of MEGI systems, with unstable network conditions and limited bandwidth, introduces communication delays that hinder real-time inference \cite{yi2025edgemoe}.
These challenges are particularly critical for complex reasoning tasks demanding sustained multi-step processing, contextual awareness, and high accuracy. To address these constraints, we propose an MoE framework augmented by CoT prompting, specifically for MEGI systems. 

\subsubsection{Design Principles}
Our design follows three core principles, implemented in the system components and workflow:
\begin{itemize}
    \item Unlike conventional token-level routing, our framework evaluates the global semantic relevance between the complete input query and expert profiles. By activating the most suitable edge device to handle the entire inference task, we minimize the communication overhead associated with frequent token exchange and ensure that domain-specific queries, such as coding or mathematics, are routed to experts with matching capabilities \cite{yi2025edgemoe}.
    
    \item Joint Communication Computation Awareness: Task scheduling simultaneously considers device compute capabilities, channel quality, transmission power, and application-specific latency and accuracy requirements, enabling end-to-end optimization. This principle is implemented via an optimization module, which guides CoT step selection, routing principle, and power control.
    
    \item Hierarchical Collaboration: The Base Station (BS) manages global scheduling and result aggregation, while edge devices execute expert networks, establishing a clear separation between control and execution. This collaboration is reflected in the modular organization of the BS Control Unit and distributed edge experts.
\end{itemize}

\subsubsection{System Components and Workflow}
As illustrated in Fig.~\ref{Fig2}, the proposed framework comprises three primary components: the BS Control Unit, distributed Edge Experts, and the integrated CoT Reasoning Module. These components execute a collaborative inference workflow optimized for resource-constrained settings.

\textbf{The BS Control Unit} serves as the central coordinator of the inference process. It initiates the end-to-end pipeline when a user submits a natural language query. To implement task-level routing, the BS first analyzes the input query to extract its global feature representation. This representation is fed into the global gating network, which computes a relevance score for each edge expert, quantifying the match between the task's domain and the experts' specialized capabilities. Based on these scores, the BS assigns the entire inference task to the top-ranked edge expert. Instead of decomposing the task into tokens for disjoint processing, the BS dispatches the complete query to the selected edge device via downlink channels. This avoids the high communication overhead associated with frequent token exchange. Transmission is power-adaptive to ensure both energy efficiency and communication reliability.

\textbf{The Edge Devices} act as distributed computation nodes that enable system scalability within the MEGI infrastructure. 
In our framework, experts are implemented as standalone small models. Each expert is hosted on an edge device and is specialized through prior SFT on domain-specific datasets.
This fine-tuning refinement enables each expert to achieve higher baseline accuracy and improved computational efficiency for its specialized inference tasks, as in Section \ref{section:fine-tuning}. It is worth noting that while this work focuses on architectural sparsity and adaptive reasoning depth, our framework differs from model compression techniques. The deployed expert models can be readily integrated with post-training quantization to further reduce memory footprint and execution latency without altering the proposed optimization logic.

\textbf{The CoT Reasoning Module} is integrated within each expert to govern the inference process. Once assigned a task, the edge device processes the input independently. For complex queries, the system uses CoT prompting to generate intermediate reasoning steps before the final answer. The number of these steps defines the reasoning depth. Increasing reasoning depth allows the model to decompose multifaceted problems into manageable sub-steps, thereby enhancing accuracy. However, this process linearly increases computational load and energy consumption \cite{xu2025towards}. Consequently, our framework treats reasoning depth as a dynamic resource parameter. The system actively balances accuracy gains against resource costs. The base station sets the optimal depth by prompting the expert to use a specific number of steps. 
Upon completion, the edge device transmits the final response, including both the reasoning steps and the answer, back to the BS for formatting and delivery.

\section{Case Study}
To validate the effectiveness of our proposed joint optimization framework for LLM reasoning in MEGI environments, we evaluated both individual model reasoning quality and system-wide efficiency in distributed settings. Our approach first establishes foundational understanding through local model deployment and then demonstrates how these insights result in system-level performance benefits.

\subsection{Local Deployment Validation}

To establish a basic understanding of reasoning quality improvements through SFT and CoT prompting, we evaluated locally deployed Qwen3 \cite{yang2025qwen3} models on a mobile edge device. To ensure domain relevance, we used the TeleQnA dataset\cite{maatouk2025teleqna}, a specialized benchmark comprised of diverse telecommunications standards and signal processing questions. We compared two lightweight model variants: \textbf{Qwen3-Base} (Qwen3-0.6B-Base) and \textbf{Qwen3-FT} (Qwen3-0.6B after fine-tuning). Both models were evaluated on TeleQnA with CoT prompting.
The experiment was performed on a device with an integrated GPU delivering 2 TFLOPS (FP16) and 16 GB of Unified Memory. We use the term edge device to describe MEGI nodes at the laptop or embedded GPU level, which are capable of running small LLMs, as distinct from low-power IoT endpoints. Therefore, while the reported trends are informative for MEGI deployment, absolute latency and power numbers may not directly transfer to all embedded platforms.

\begin{figure*}[t]
\centering{\includegraphics[width=0.96\textwidth]{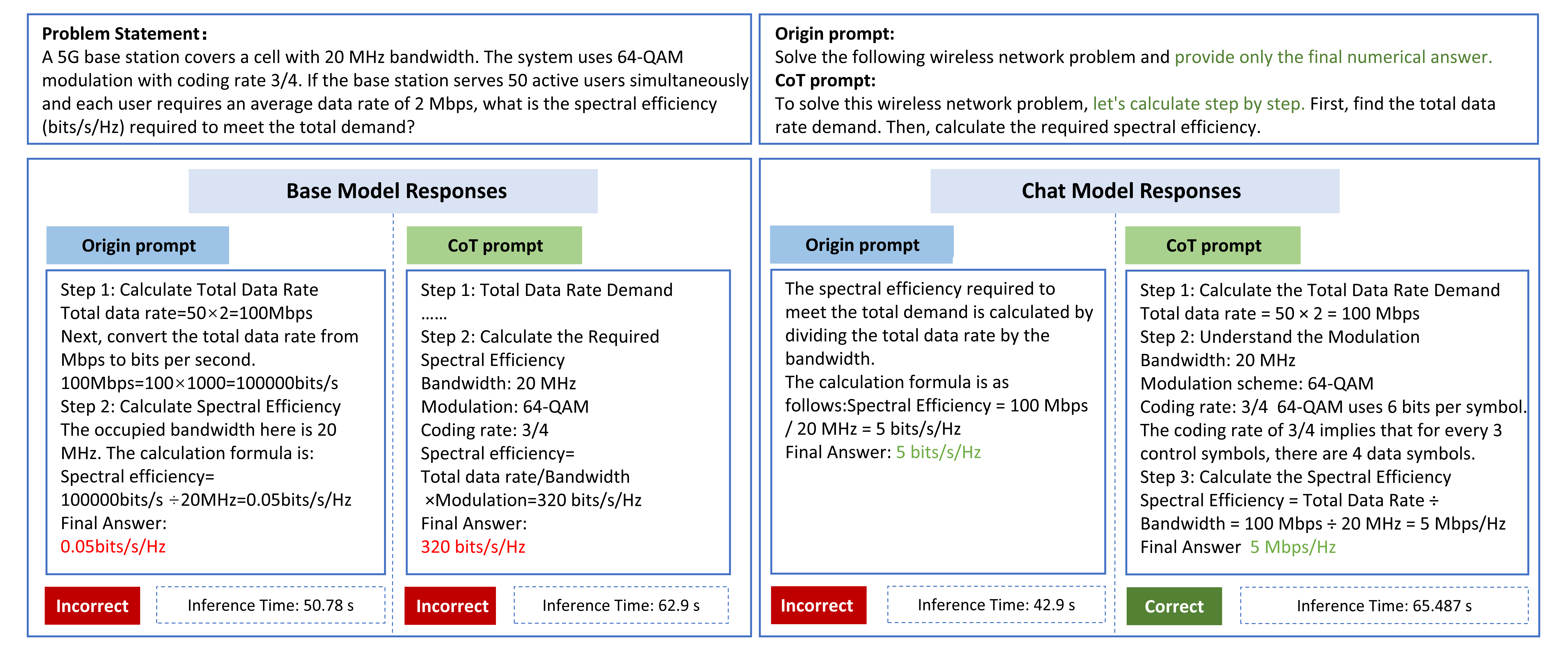}}
\caption{Summary of output correctness and inference time for \textbf{Qwen3-FT} and \textbf{Qwen3-Base} with and without CoT prompts on a mobile edge device.}
\label{Fig4}
\end{figure*}

Fig. \ref{Fig4} shows the Qwen3-FT model is more accurate and faster than the base model. This highlights the importance of model selection in resource-constrained environments. While SFT models require initial training overhead, they provide substantial operational efficiency gains that justify the investment for deployment scenarios with sustained usage patterns. Regarding inference strategies, CoT prompting improves interpretability but inherently increases generation latency, because decoding latency scales linearly with the number of generated tokens. Additionally, CoT sometimes introduces extra errors in the base model. Consequently, the Qwen3-Base model with direct prompting achieves an optimal balance between accuracy and speed. In contrast, our observations show the fine-tuned model produces longer, more detailed reasoning traces under CoT prompts. While providing deeper insights, this can make the fine-tuned model slower than the Base variant in actual wall-clock time.
To ensure the statistical reliability, we tested the models on TeleQnA dataset \cite{maatouk2025teleqna}, which comprises 1000 diverse questions.
The specialized Qwen3-FT model achieves an accuracy of 49.7\% with CoT, outperforming the base model (28.9\%). An accuracy of 49.7\% may appear modest in isolation, but TeleQnA is a difficult benchmark for evaluating fine-grained telecommunications expertise. Our focus is evaluating the accuracy-efficiency trade-off for a lightweight model on MEGI nodes, rather than targeting state-of-the-art results.
These empirical results guide our distributed MEGI framework design. First, the efficiency of SFT models justifies deploying specialized experts within the MoE layer to overcome edge constraints. Second, the CoT accuracy-latency trade-off requires dynamic control. We treat reasoning depth as an adaptive variable. The system increases depth for complex queries under good channel conditions and reduces it during network congestion to meet strict latency limits.

\subsection{System-Level Performance Evaluation}
Building on insights from the local deployment validation, we conducted system-level evaluations to examine how individual reasoning improvements translate into broader performance gains in distributed MEGI environments.
Following the framework presented in Section \ref{section:system_model}, our experimental evaluation was conducted within a simulated MEGI environment covering a $1000m \times 1000m$ area, where a central BS coordinates 15 distributed edge devices. The experiment was conducted over 1000 time slots.  Tasks arrive at the BS following a Poisson process, reflecting the stochastic nature of user requests in cellular networks. The tasks are diverse, requiring different domain expertise (e.g., logic, coding, or general QA). To simulate realistic edge constraints, we adopted standard cellular specifications with a 23 dBm transmission power limit and 20 MHz bandwidth, utilizing a channel model that accounts for path loss and shadowing. Each edge expert is modeled as a Qwen3-FT instance (occupying $\approx$ 1.2 GB memory) running on a node with 16 GB Unified Memory and 2 TFLOPS compute capability.

We formulated a joint optimization problem to minimize system energy while maximizing the satisfaction of accuracy and latency requirements (thresholds set to [0.75, 0.85] and [50s, 60s]). These latency thresholds reflect latency-tolerant MEGI services such as network troubleshooting, configuration recommendation, and fault diagnosis with human-in-the-loop decision-making, where a response within about one minute is often acceptable. For stricter real-time tasks, tightening the latency budget prompts the adaptive mechanism to automatically reduce CoT depth. Energy consumption and latency represent the combined cost of communication and model computation. The total latency consists of transmission delay, queuing delay (waiting time when the expert is busy), and processing delay (which scales linearly with the CoT reasoning depth). To solve this, we employed Distributed Proximal Policy Optimization (DPPO). By observing states like queue length and channel quality, the DPPO agent jointly optimizes: (1) \textit{Expert Selection} for domain matching; (2) \textit{Power Control} for link optimization; and (3) \textit{Reasoning Depth} to balance accuracy against computational/communication costs. Finally, the reward function incentivizes high accuracy while penalizing the extra energy and latency costs caused by additional reasoning tokens.

We evaluate four distinct system configurations to isolate the contributions of different architectural and reasoning components: (1) Dense model baseline - traditional centralized LLM deployment at the BS without CoT; (2) MoE without CoT - distributed MoE architecture handling tasks via zero-shot inference, utilizing static power control; (3) MoE with Fixed CoT - a heuristic approach employing predetermined CoT depth, uniform expert routing, and static transmission power; and (4) MoE with Dynamic CoT - our proposed framework with adaptive CoT depth selection and joint resource optimization via DPPO. 
Since no prior work in wireless networks has explicitly implemented or optimized reasoning in LLM-based inference, there are no directly comparable baselines. Therefore, our comparison focuses on controlled architectural and reasoning variations within our framework. 
The performance is measured using three key metrics:
total energy consumption, cumulative energy for all inference requests; accuracy satisfaction rate, percentage of tasks exceeding a set reasoning quality threshold aggregated over the 1000 stochastic time slots; and latency satisfaction rate, percentage of tasks completed within specified latency constraints.
% This systematic comparison allows us to quantify the individual and combined effects of architectural sparsity (MoE), structured reasoning (CoT), and adaptive optimization (dynamic depth selection) on system performance.
\begin{figure}[t]
\centering{\includegraphics[width=0.46\textwidth]{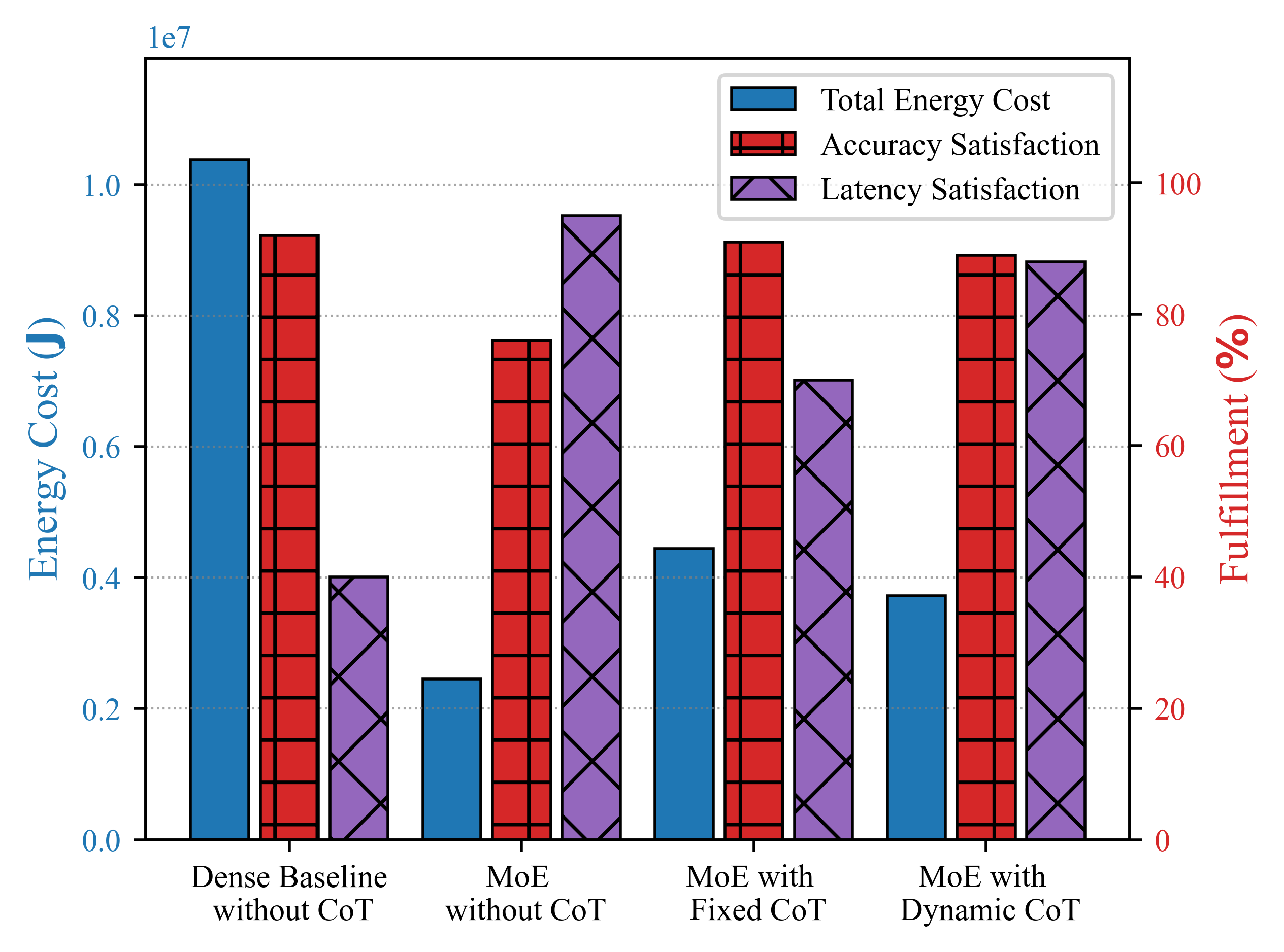}}
\caption{Experimental results of comparing energy consumption, accuracy, and latency satisfaction for four inference schemes.}
\label{Fig3}
\end{figure}

Fig.~\ref{Fig3} shows our joint optimization approach consistently outperforms baseline and partial schemes. The distributed MoE architecture improves scalability through efficient resource allocation. Specifically, it reduces total energy consumption by routing routine requests to lightweight experts and reserving heavy computation for complex queries. Furthermore, CoT prompting further enhances accuracy satisfaction. Meanwhile, adaptive scheduling and expert allocation ensure that latency constraints are met in over 90\% of the cases. This validates that the synergistic combination of scalability enhancement through MoE and reasoning enhancement through adaptive CoT enables effective deployment of sophisticated reasoning capabilities in resource-constrained MEGI environments.
Regarding implementation complexity, the Heuristic Baseline offers the lowest deployment overhead ($O(1)$ complexity) but lacks adaptability. While our proposed scheme requires offline DPPO training, its online inference complexity is negligible, enabling real-time responsiveness. To ensure the real-time feasibility of our framework, we measured the execution time of the DPPO agent. The average inference latency is approximately 150 ms per decision, which is negligible compared to the task execution latency, confirming that the optimization overhead does not create a bottleneck.

\subsection{Discussion and Analysis}
The experiments validate our framework's design at both individual model performance and system-wide optimization. Local tests highlight the trade-off between efficiency and reasoning quality. Specifically, SFT models improve performance, but CoT prompting adds computational costs. These findings directly guided our distributed architecture and the use of adaptive CoT depth. 
The system-level evaluation confirms that these individual model improvements translate to significant system-wide benefits. The distributed MoE architecture effectively mitigates computational bottlenecks in dense LLM deployments, while the adaptive CoT mechanism balances energy efficiency, reasoning accuracy, and response time through intelligent depth selection. 
% The DPPO optimization exhibits superior convergence in the complex multi-objective optimization space, validating its suitability for MEGI scenarios where traditional approaches are intractable.	

\section{Future Directions}
\textbf{Security and Privacy:}
Securing distributed LLM reasoning remains a critical challenge. Sensitive data exchanged between mobile devices and edge servers is prone to tampering and unauthorized access, risking both reasoning accuracy and user privacy. The distributed MoE architecture introduces additional vulnerabilities, where malicious nodes could extract parameters or alter outputs. Future work should explore privacy-preserving techniques-such as differential privacy, secure multi-party computation, and homomorphic encryption-adapted for distributed reasoning, along with lightweight authentication and verification mechanisms.

\textbf{Multi-Modal Reasoning:}
Most LLM reasoning frameworks are limited to text, hindering their utility in real-world scenarios requiring multimodal understanding. Future MEGI systems should support reasoning across visual, auditory, and sensor data. This calls for lightweight fusion architectures capable of coherent cross-modal reasoning, optimized for heterogeneous edge devices and dynamic resource availability.

\textbf{Decentralized Collaboration:}
Centralized base station coordination creates bottlenecks and limits scalability. Edge nodes often operate independently, missing opportunities for collaborative reasoning. Future systems should support decentralized coalitions of edge devices that coordinate via distributed consensus, share intermediate reasoning states, and adaptively balance workloads based on device heterogeneity and network conditions. Such collaboration can improve fault tolerance and enable richer, distributed reasoning capabilities.

\section{Conclusion}
In this article, we have explored the integration of LLM reasoning into MEGI systems. We have provided a comprehensive overview of reasoning enhancement techniques across the pre-training, fine-tuning, and inference stages, establishing a solid theoretical foundation for reasoning-aware system design at the edge. We have proposed a joint optimization framework that combines a distributed MoE architecture with adaptive CoT prompting. This framework improves energy efficiency, preserves reasoning quality, and reduces latency through a DPPO-based optimization approach. Finally, we have validated the practical effectiveness of our framework through case studies and have outlined future research directions to further advance LLM reasoning for MEGI applications.

\bibliographystyle{IEEEtran}
\bibliography{mylib}%,IEEEexample}

\end{CJK}
\end{document}